\documentclass{article}
\usepackage{ijcai21}

\usepackage{times}
\usepackage{soul}
\usepackage{url}
\usepackage[hidelinks]{hyperref}
\usepackage[utf8]{inputenc}
\usepackage[small]{caption}
\usepackage{graphicx}
\usepackage{amsmath}
\usepackage{amsthm}
\usepackage{booktabs}
\usepackage{amsmath}
\usepackage{amsfonts}
\usepackage[ruled,linesnumbered]{algorithm2e}
\usepackage[switch]{lineno}
\usepackage{subcaption}
\usepackage{multirow}
\usepackage{appendix}

\title{Solving Large-Scale Extensive-Form Network Security Games via  Neural Fictitious Self-Play}

\author{
Wanqi Xue$^1$
\and
Youzhi Zhang$^2$\and
Shuxin Li$^1$\and
Xinrun Wang$^1$\and
Bo An$^1$\And
Chai Kiat Yeo$^1$
\affiliations
$^1$School of Computer Science and Engineering, Nanyang Technological University, Singapore\\
$^2$Department of Computer Science, Dartmouth College, USA\\
\emails
wanqi001@e.ntu.edu.sg,
youzhi.zhang@dartmouth.edu,
\{shuxin.li, xinrun.wang, boan, asckyeo\}@ntu.edu.sg
}

\usepackage{bm}

\begin{document}

\maketitle

\begin{abstract}

Securing networked infrastructures is important in the real world. The problem of deploying security resources to protect against an attacker in networked domains can be modeled as Network Security Games (NSGs). Unfortunately, existing approaches, including the deep learning-based approaches, are inefficient to solve large-scale extensive-form NSGs. In this paper, we propose a novel learning paradigm, NSG-NFSP, to solve large-scale extensive-form NSGs based on Neural Fictitious Self-Play (NFSP). Our main contributions include: i) reforming the best response (BR) policy network in NFSP to be a mapping from action-state pair to action-value, to make the calculation of BR possible in NSGs; ii) converting the average policy network of an NFSP agent into a metric-based classifier, helping the agent to assign distributions only on legal actions rather than all actions; iii) enabling NFSP with high-level actions, which can benefit training efficiency and stability in NSGs; and iv) leveraging information contained in graphs of NSGs by learning efficient graph node embeddings. Our algorithm significantly outperforms state-of-the-art algorithms in both scalability and solution quality. 
\end{abstract}

\section{Introduction}
How to secure networked infrastructures, e.g., urban city networks, transportation networks, and web networks, has received extensive attention \cite{Jain11,yz17,yz19,okamoto2012solving}. The problem of deploying a limited number of security resources (controlled by the defender) to protect against an attacker in networked domains can be modeled as Network Security Games (NSGs). 
We consider a realistic game setting where the players interact sequentially (extensive-form) and the defender makes decisions based on real-time information about the attacker \cite{yz19}.
The objective of NSGs is to find a Nash Equilibrium (NE) policy for the defender.  Traditionally, the defender's policy is computed by programming-based NE-solving techniques, e.g., the incremental strategy generation algorithms \cite{doubleoracle,yz19}, which start from a restricted game and iteratively expand it until convergence.
One important requirement of these approaches is that all of the attacking paths  are enumerable, which is to ensure that there is at least a terminal state in the restricted game for each attacking path to make the incremental strategy generation algorithm converge. However, in large-scale NSGs, e.g., real-world road networks, where the number of attacking paths are prohibitively large, programming-based NE-solving approaches tend to lose effectiveness. For example, in the real world, the number of possible attacking  paths could be more than $6.6^{18}$ \cite{Jain11}, which will make it impossible to  enumerate all of them  due to the limited memory.

Recently, there has been an increasing interest in combining Deep Learning (DL) with game theory for finding NE \cite{nfsp,psro,deepcfr}. DL-based NE-solving algorithms use Deep Neural Networks (DNNs) to learn states-to-actions mappings for approximating strategies, counterfactual regrets, etc. They usually execute in a sampling style and are able to capture the structure of underlying enormous state spaces by leveraging strong representation ability of DNNs, making them potential for solving large-scale and complex real-life problems.

Unfortunately, existing DL-based NE-solving algorithms are unable to solve large NSGs.
In NSGs, players occupy nodes of graphs, e.g., road networks, and can only move to their adjacency nodes (legal actions) at each step. A consequence is that legal actions change with players' current positions or states. To approximate states-to-actions mappings, a naive implementation is to set the output dimension of DNNs equal to the maximum number of legal actions, with each output corresponding to one legal action though the action changes with states. However, this naive setting yields poor results in practice because each output of DNNs has no consistent semantics \cite{growingactionspace}. On the other hand, it is infeasible to set the output dimension of DNNs equal to the number of all actions and use masks to filter out all illegal actions at each state. The reason is that, in NSGs, the defender's action space is prohibitively large because it is a combination of all sub-action spaces of the defender's security resources. For example, when there are 
four security resources deployed on a road network with one hundred nodes, the defender's action set has $100^4$ elements. Obviously, we cannot define the output of DNNs at such a scale.

In this paper, we propose a novel learning paradigm, NSG-NFSP, for approximating an NE policy in large-scale extensive-form NSGs. The method is based on Neural Fictitious Self-Play (NFSP), which intrinsically ensures its convergence.
Our main contributions are fourfold. Firstly, we propose to train the best response (BR) policy network in NFSP to be a mapping from action-state pair to action-value, which avoids the aforementioned unachievable requirement where the output of DNN must cover the overall action spaces of NSGs.
Secondly, we convert the average policy network into a metric-based classifier, helping an NFSP agent to assign distributions only on legal actions rather than all actions. Thirdly, we propose a framework to enable NFSP with high-level actions, which can enhance training efficiency and stability in NSGs.
Finally, we propose to learn efficient graph node embeddings by \textit{node2vec}, to leverage information contained in the graphs of NSGs.
We conduct experiments in NSGs played on synthetic networks and real-world road networks. Our algorithm significantly outperforms state-of-the-art algorithms in both scalability and solution quality.

\section{Preliminaries and Related Works}
  \subsection{Network Security Games}
Network Security Games (NSGs) are proposed to model the problem of deploying a limited number of security resources to protect against an adaptive attacker in networked domains \cite{Jain11}. For example, the police department distributes collaborating police officers to prevent a criminal from escaping or attacking in urban cities \cite{yz17,Jain11}. An NSG is played on a graph $G=(V,E)$ which consists of a set of edges $E$ and a set of nodes $V$. 
There are two players, the defender and the attacker. The attacker, starting from one of the source nodes $v^{att}_0 \in V_s \subset V$, tries to reach one of the target nodes $\chi \in V_t \subset V$ within a fixed time horizon $T$\footnote{The target nodes represent destinations to be attacked or exits to escape.}. The defender controls $m$ security resources and dynamically allocates them to catch the attacker before he reaches any of the targets. We assume that the defender can observe the real-time location of the attacker, with the help of advanced tracking technologies such as GPS, but the attacker can only see the initial locations of the security resources \cite{yz19}. We model NSGs as extensive-form games, where players make decisions sequentially.

At time step $t$, the attacker's state $s_t^{att}$ is a sequence of nodes he has visited, i.e., $s_t^{att}=\langle v_0^{att},v_1^{att},\dots, v_t^{att} \rangle$. $\mathcal{A}_{att}=V$ is the set of attacker actions and $\mathcal{A}_{att}(s_t^{att})=\{v_{t+1}^{att}|(v_{t}^{att},v_{t+1}^{att})\in E\}$ is the set of legal attacker actions at $s_t^{att}$. For the defender, its state $s_t^{def}$ consists of the attacker's action history (state) and its resources' current locations, i.e., $s_t^{def}=\langle s_t^{att},
l_t^{def}\rangle$ where $l_t^{def}=\langle v_t^0,\dots, v_t^{m-1} \rangle$.  $l_t^{def}$ is adjacent to resource location $\langle v_{t+1}^0,\dots, v_{t+1}^{m-1} \rangle$ if $(v_{t}^R,v_{t+1}^R)\in E,\forall R \in \{0,\dots,m-1\}$. We denote $Adj(l_t^{def})$ as the set of all resource locations which are adjacent to $l_t^{def}$. With this concept, we can define the set of legal defender actions at $s_t^{def}$ as $\mathcal{A}_{def}(s_t^{def})=Adj(l_t^{def})$ and
$\mathcal{A}_{def}=V^m$ is the set of defender actions. For both the attacker and the defender, illegal actions at state $s$ are those actions in $\mathcal{A}$ but not in $\mathcal{A}(s)$. 
A policy $\pi(s)=\Delta(\mathcal{A}(s))$ describes a player's behavior, where $\Delta(\cdot)$ represents a probability distribution. Both players act simultaneously at each step by sampling actions from their policies: $a\sim \pi(s)$. The attacker is caught if he and at least one of the security resources are in the same node at the same time. A game ends when the attacker either reaches any of the targets within the maximum allowed time or is caught. 
If the defender successfully protects the targets within the time horzion $T$, she will be awarded with a positive unit utility (an end-game reward) $u_{def}=1$. Otherwise, no award will be given to the defender. The game is zero-sum, so $u_{att}=-u_{def}$. 
The worst-case defender utility $\mathcal{E}_{def}(\pi_{def})$ is the expected payoff for the defender (with policy $\pi_{def}$) given that the attacker best responds to it. Formally,
 $\mathcal{E}_{def}(\pi_{def})=\min_{\pi_{att}} \mathbb{E} \big[u_{def}|\pi_{def},\pi_{att}\big]$.
$\pi_{def}^*$ is optimal if 
$\pi_{def}^* \in \arg\max_{\pi_{def}}\mathcal{E}_{def}(\pi_{def})$.
The optimality for the attacker is defined similarly.
A Nash Equilibrium (NE) is reached if and only if both the defender and the attacker perform the optimal policy. In NSGs, the optimization objective is to learn an NE policy for the defender.

\subsection{Neural Fictitious Self-Play}
\label{nfsp}

\begin{figure}[t]
    \centering
    \includegraphics[width=0.48\textwidth]{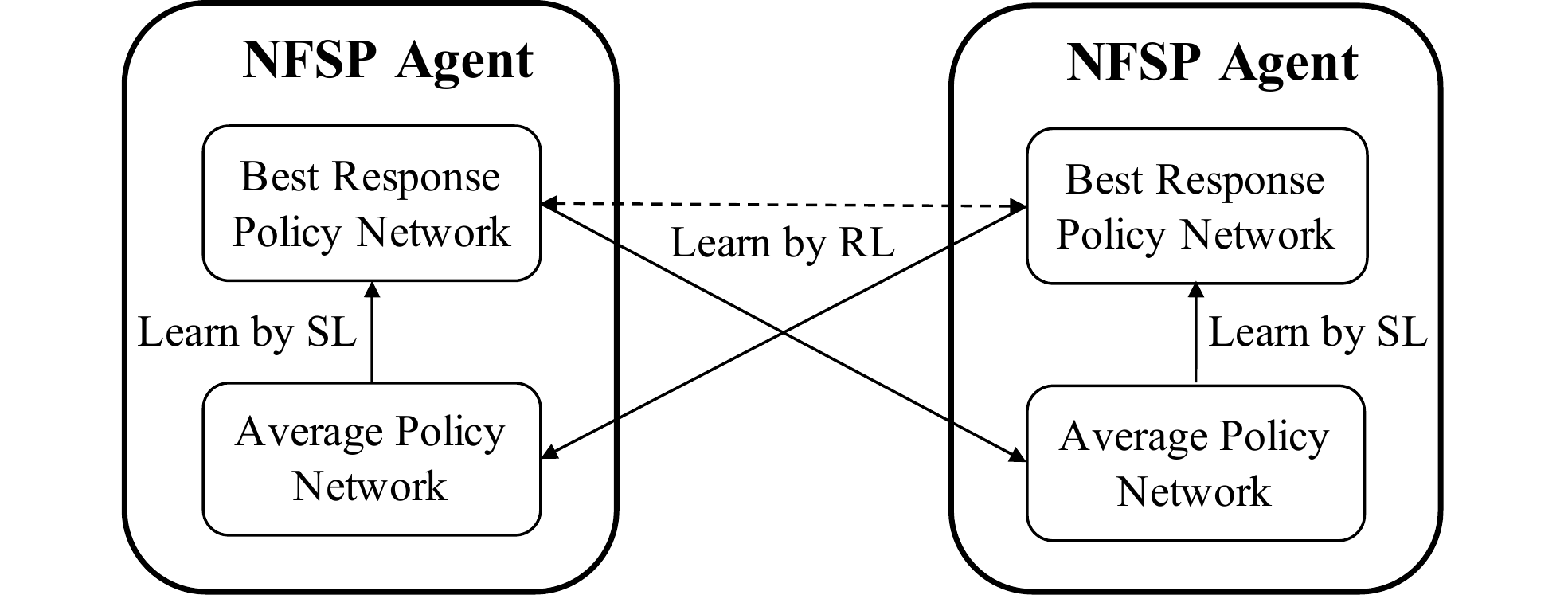}
    \caption{The NFSP framework.}
    \label{nfsp_frame}
\end{figure}

\textbf{Fictitious play} (FP) \cite{fp} is a game-theoretic algorithm for learning NE from self-play. In FP, each agent plays with its opponent's past average policy and best responds against it. 
\textbf{Fictitious Self-Play (FSP)} \cite{xfp} extends FP from normal form to extensive form and realizes it in a sampling and machine learning style. 
\textbf{Neural Fictitious Self-Play (NFSP)} \cite{nfsp} combines FSP with neural network function approximation. 
As in Figure \ref{nfsp_frame}, each NFSP agent consists of two neural networks, i.e., the best response (BR) policy network and the average policy network. The BR policy network is trained by reinforcement learning (RL) algorithms, e.g., DQN \cite{dqn}, to maximize the expected total rewards. It considers the opponent as part of the environment.
The average policy network is trained to approximate the past average behaviours of the BR policy network by supervised learning (SL). It outputs the probabilities of actions chosen, historically, by the BR policy network.
Each NFSP agent behaves according to a mixture of its BR policy and average policy (with a mixing constant $\eta$ which is called anticipatory parameter).

Most applications of NFSP are limited in domains with small discrete action spaces. Despite this, applying NFSP to other types of action spaces has received extensive attention. OptGradFP \cite{cfsp1} firstly introduces fictitious play to continuous action spaces. It applies policy gradient algorithm over policy network which predicts parameters of continuous action distributions. DeepFP \cite{cfsp2} generalizes OptGradFP by using flexible implicit density approximators. Currently, applying NFSP to games like NSGs, whose action spaces are large and legal actions vary significantly with states, remains unexplored. The main challenge is that the output of the two DNNs in NFSP cannot cover all actions, and if just covering legal actions, they will lack consistent semantics. 
With many recent works focusing on adapting deep RL to large discrete action spaces~\cite{acl,ar}, we propose our solution for the BR policy network based on DRRN \cite{acl}. It models Q-values as an inner product of state-action representation pairs. DRRN is designed for natural languages domain. We adapt it to make it suitable for NSGs whose actions are defined on graph nodes. For the average policy network, our solution is inspired by metric-based few-shot learning \cite{protonet}. We propose to address the problem by transforming actions and states to a space where the probabilities of actions can be determined via comparing some metrics, e.g., cosine similarity. Further discussions about related works are provided in the appendix of the full version.

\section{Methodology}
In this section, we introduce a novel learning paradigm, NSG-NFSP, for solving large-scale NSGs. Note that despite the method being proposed for solving NSGs, the basic ideas can be easily applied to other games, especially those whose legal action spaces vary significantly with states. We provide the overall algorithm in the appendix.

\begin{figure}[t]
    \centering
    \includegraphics[width=0.48\textwidth]{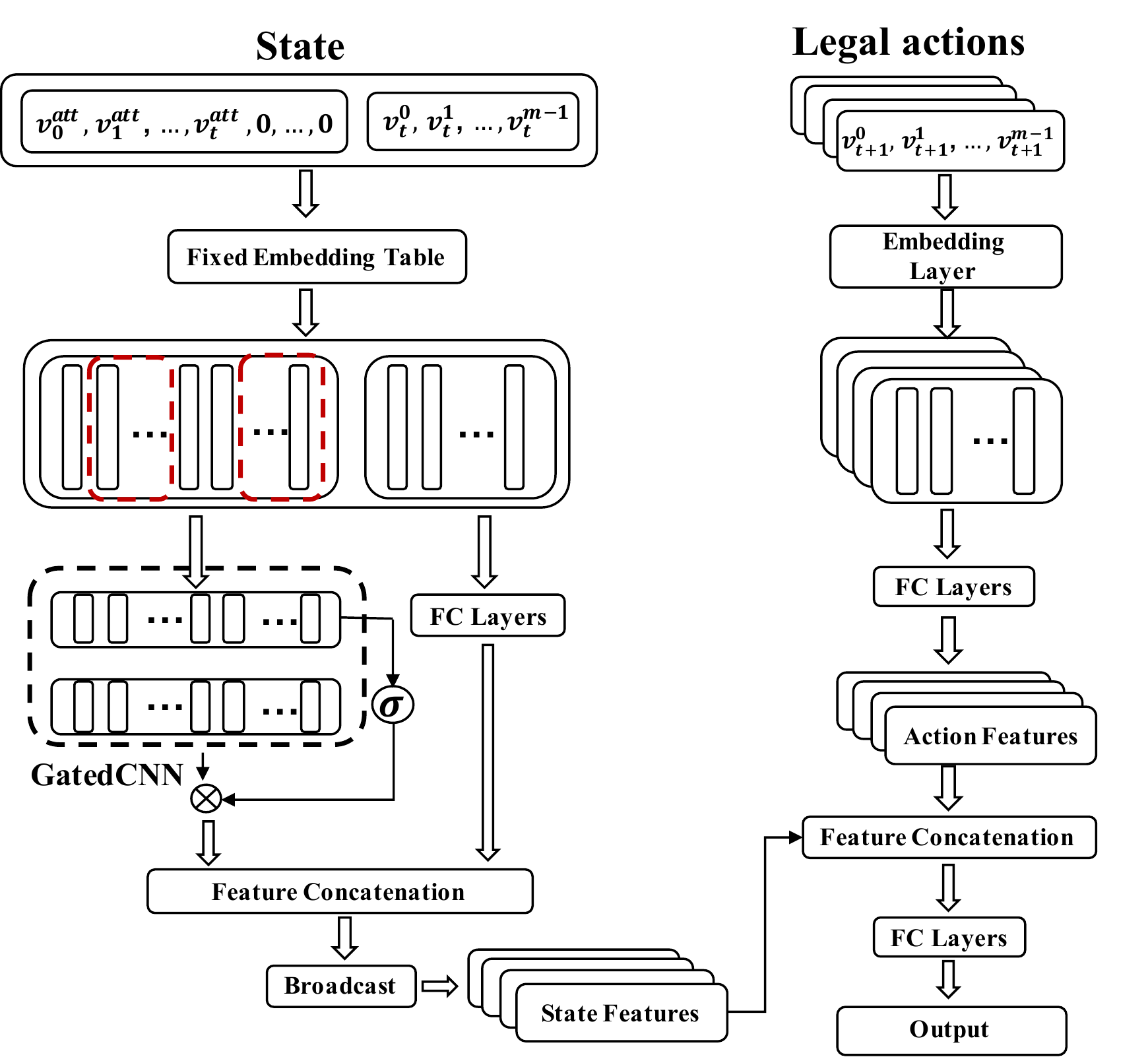}
    \caption{The network structure of the defender (the two red boxes indicate the convolutional blocks in GatedCNN).}
    \label{drrn}
\end{figure}
\subsection{Approximating Best Response Policy}
\label{abrp}
It is essential to properly approximate the BR policy in NFSP because the final (average) policy is supervisedly trained from the behavior of the BR policy network. The BR policy network in the vanilla NFSP algorithm learns a mapping from states to action-values, which internally requires the DNN's outputs to cover all possible actions. However, for games like NSGs, it is impossible to meet this requirement because the overall action space is enormous. To address the problem, we propose to convert the BR policy network to be a mapping from state-action pairs to Q-values. Concretely, we use an action representation network and a state representation network, parameterized by $\theta^Q_{\alpha}$ and $\theta^Q_{\beta}$, to extract features from each legal action and state respectively, generating feature vectors $h_a$ and $h_s$. The extracted features $h_a$ and $h_s$ are concatenated and sent to a fully connected network $f(h_a,h_s;\theta^Q_{\gamma})$ with parameters $\theta^Q_{\gamma}$ to predict the action-value. We denote $\theta^Q=\{\theta^Q_{\alpha},\theta^Q_{\beta},\theta^Q_{\gamma}\}$ 
as the parameters of the BR policy network. During training, an agent stores its experienced transition tuples, $(s,a,r,s',\mathcal{A}(s'))$, in a replay buffer $\mathcal{M}_{RL}$, where $r \sim R(\cdot|s,a)$ (reward function) is the immediate reward and $s' \sim P(\cdot|s,a)$ (transition function) is the next state.
The BR network parameters $\theta^Q$ are optimized by minimizing the loss:
\begin{equation}
    \mathcal{L}(\theta^Q)=\mathbb{E}_{s,a,r,s'}\Big[(r+\max_{a'\in\mathcal{A}(s')} Q(s',a';\theta^{Q'})-Q(s,a;\theta^Q))^2\Big]
 \label{loss_rl}
\end{equation}
where $\theta^{Q'}$ denotes the parameters of the target network. $\theta^{Q'}$ is periodically copied from $\theta^{Q}$, while in other cases it is frozen to improve the stability of training.

Figure \ref{drrn} presents the overall network architecture, NSG-BR, for the defender. We can design the network for the attacker similarly. Since the elements of states and actions are graph nodes, we firstly embed those graph nodes before they can be fed into neural networks. For the action representation network, we use a learnable embedding layer. For the state representation network, we pre-compute the embeddings (the approach is introduced in Section \ref{egne}).
After embedding graph nodes, we need to extract features from the attacker's history. Taking into account the speed and effectiveness, we apply a structure similar to  GatedCNN \cite{gatedcnn} to process these sequential data.
Specifically, the sequential data padded to the maximum length is fed into two separate convolutional blocks which have identical structures. The output of one block is activated by the sigmoid function $\sigma(x)=\frac{1}{1+e^{-x}}$, and the result serves as the gate to control the output of the other convolutional block. After extracting features for a state, the state feature vector is duplicated (broadcast) $|\mathcal{A}(s)|$ times (number of legal actions at the state) and concatenated with legal action features. Then the state-action pairs' features are passed to several fully connected layers to predict the final state-action values.

\subsection{Approximating Average Policy}
\label{aap}
The average policy network in the vanilla NFSP algorithm is a classifier-like network whose output scales linearly with the cardinality of action set $\mathcal{A}$. In NSGs, assigning distributions over legal action set $\mathcal{A}(s)$, rather than the whole action set, is preferred. We propose to convert the average policy network into a metric-based classifier, transforming states and legal actions to a space where they can be compared by some metrics.
We use fully-connected layers to learn the metric. The reason is that i) learnable metric is more representative compared to heuristic metric; and ii) we can reuse the network architecture as depicted in Figure \ref{drrn}. Similar to NSG-BR, the average policy network, NSG-AVG, has three parts, i.e., the action representation network, the state representation network and the metric network.
We denote $\theta^\Pi=\{\theta^\Pi_{\alpha},\theta^\Pi_{\beta},\theta^\Pi_{\gamma}\}$ to be parameters of NSG-AVG, where $\theta^\Pi_{\alpha},\theta^\Pi_{\beta},\theta^\Pi_{\gamma}$ are the parameters of its three sub-networks respectively. NSG-AVG works similarly to NSG-BR, the main difference being that NSG-AVG assigns probabilities to legal actions rather than Q-values.
 Let $\Pi(\mathcal{A}(s)|s)$ be the output of NSG-AVG and  $\hat{\Pi}(\mathcal{A}(s)|s)$ be the ground truth average BR behaviours. We measure the difference between these two distributions, $\Pi(\mathcal{A}(s)|s)$ and $\hat{\Pi}(\mathcal{A}(s)|s)$, by the expected Kullback-Leibler (KL) divergence. Then the parameters $\theta^\Pi$ of NSG-AVG can be optimized by minimizing the difference:
\begin{equation}
\vspace{-0.05cm}
\begin{split}
\mathcal{L}(\theta^\Pi) & = -\mathbb{E}_s \Big[\sum_{a\in \mathcal{A}(s)}\hat{\Pi}(a|s)\log \Big(\frac{\Pi(a|s)}{\hat{\Pi}(a|s)}\Big) \Big] \\
 & = -\mathbb{E}_{s,a} \Big[\log \Big(\frac{\Pi(a|s)}{\hat{\Pi}(a|s)}\Big) \Big]
\end{split}
\label{slloss}
\end{equation}
The denominator in Eq.(\ref{slloss}) can be omitted since it does not depend on $\theta^\Pi$. Then the optimization objective becomes minimizing the loss function:
\begin{equation}
  \mathcal{L}(\theta^\Pi) =  -\mathbb{E}_{s,a} \Big[\log \Big(\Pi(a|s)\Big) \Big]
 \label{loss_sl}
\end{equation}
An agent records its BR behaviours, i.e., $(s,a)$, in a reservoir buffer $\mathcal{M}_{SL}$ which serves as an expanding dataset. It fits the dataset by applying gradient descent. According to the theoretical convergence of NFSP \cite{nfsp}, the average policy network approximates an NE policy.

\begin{figure}
    \centering
    \includegraphics[width=0.4\textwidth]{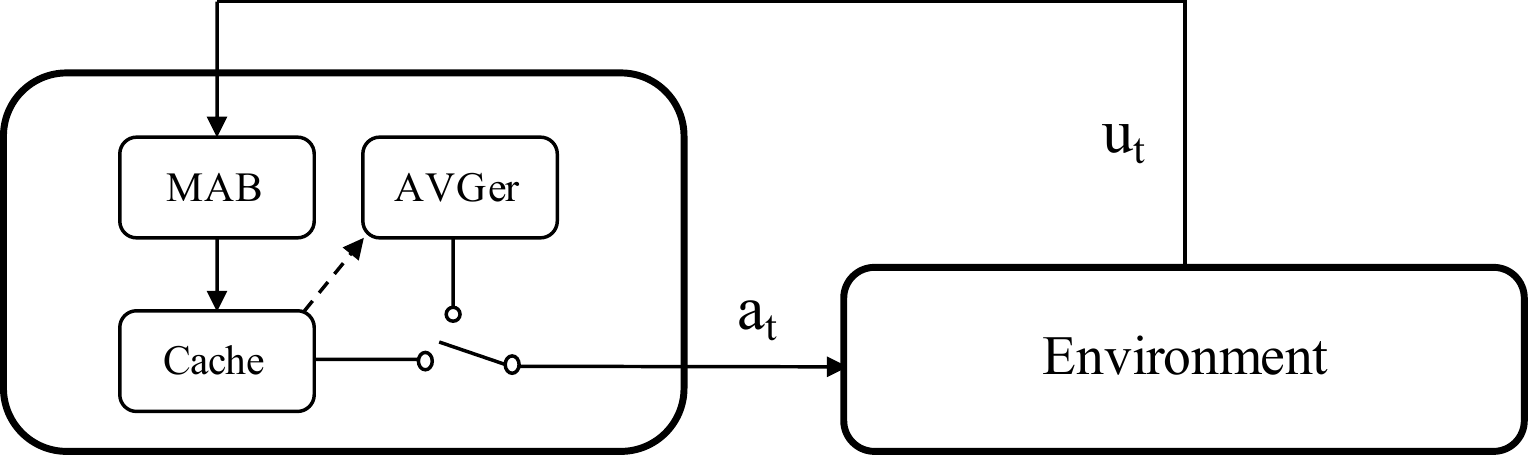}
    \caption{The architecture of NFSP with high-level actions.}
    \label{atk_frame}
\end{figure}

\subsection{Enabling NFSP with High-Level Actions}
\label{mtap}
During training, it usually takes many episodes for the BR policy network to approximate the BR policy. For example, in NSGs, the NFSP attacker will do a lot of unnecessary explorations in invalid paths, i.e., the paths which cannot reach any of the targets, before finding the optimal path. This will lead to training inefficiency and instability because i) other agents will play against this weak opponent for a long period; and ii) behaviours of non-BR policy will be used for training the average policy network. The problems can be mitigated if an NFSP agent makes decisions on high-level actions (HLA). For instance, we can force the NFSP attacker in NSGs to make decisions on valid paths or source-target pairs (high-level actions) rather than the next-step location. The idea of HLA is similar to action abstractions \cite{ab1,ab2} and options in hierarchical RL \cite{op}.

To extend the NFSP framework so that an agent can decide on High-Level Actions (NFSP-HLA),
we propose to use Multi-Armed Bandit (MAB), a widely used approach to optimize decisions between multiple options (actions), to model the BR policy for an NFSP-HLA agent. Each option of the MAB corresponds to a high-level action. We use a first-in-first-out (FIFO) buffer with length $k$ to record the most recent $k$ game results (utilities). The estimated action value for a high-level action $\zeta$ after $n$ episodes becomes:
\begin{equation}
  \hat{Q}_n(\zeta) = \left.
    \frac{\sum_{j=\tau}^n u_{j} \mathbb{I}[\zeta_j = \zeta]}{\sum_{j=\tau}^n  \mathbb{I}[\zeta_j = \zeta]}
  \right.
  \label{Q-MAB}
\end{equation}
where  $\tau=\max(n-k,1)$, $u_j$ is the player's utility for the $j$-th episode, and $\mathbb{I}$ is binary indicator function. We design two auxiliary modules to fit the MAB best responsor into the framework of NFSP: i) the Averager (AVGer) module, which is used to measure the average policy, by counting the frequency of each high-level action; and ii) the Cache module, which is to temporarily store behaviours of the MAB. Data stored in the Cache is used to update the AVGer.

\paragraph{Learning Process } As in Figure \ref{atk_frame}, before each episode, an NFSP-HLA agent samples its behaviour pattern, acting as either the MAB (with probability $\eta$) or the AVGer (with probability $1- \eta$). If acting as the MAB, the agent chooses the high-level action with the largest estimated value and stores the high-level action in the Cache. Otherwise, the agent samples an high-level action according to the distribution in the AVGer. After confirming the high-level action, the agent interacts with the environment (containing the opponents) and receives an utility at the end of a game. Then the utility is used for training the MAB in accordance with Eq. (\ref{Q-MAB}). The Cache pours its records into the AVGer in a fixed frequency, after which it clears itself.  By using the Cache, we can avoid rapid changes in the AVGer, thus reducing instability.

\paragraph{Additional Exploration } An NFSP-HLA agent does exploration by acting as the AVGer (not the MAB). Such mechanism may lead to suboptimality when some actions are dominated. For example, if the AVGer explores some good actions, the MAB is likely to choose them because of their high estimated values. Behaviours of the MAB will be 
recorded by the AVGer, further increasing these actions’ occurrence frequency (selected by the AVGer). This may result in some actions appearing rarely, and the MAB cannot precisely estimate those actions' values. To overcome this, we design additional exploration for 
NFSP-HLA agent: if the agent does not act as the MAB, it will perform additional sampling to decide whether to explore or not. If the sampling result indicates exploration, the agent will act randomly and the transitions for this episode will not be recorded by the opponent; Otherwise, the agents will interact normally. Additional exploration confirms that each high-level action occurs enough times that the MAB can conduct action-value estimation.

\subsection{Efficient Graph Node Embeddings}
\label{egne}
To leverage information, e.g., adjacency and connectivity, contained in graphs of NSGs, we propose to use \textit{node2vec} \cite{node2vec}, a semi-supervised learning algorithm, to learn representations for nodes. 
Concretely, we learn a mapping function to transform nodes to low-dimensional space of features which maximizes the likelihood of preserving graph structures. Let $g:V \rightarrow \mathbb{R}^D$ be the mapping function, we need to optimize:
\begin{equation}
    \max_g \sum_{v\in V}\log \textit{Prob}(N_O(v)|g(v))
\end{equation} where
$N_O(\cdot)$ is a randomized procedure that samples many different neighborhoods of a given node. Concretely, the procedure $N_O(\cdot)$ is parameterized by a sampling strategy $O$:
\begin{equation}
    O(v_t=x|v_{t-1}=u)=\begin{cases}
    \frac{U(u,x)}{Z} &\text{if } (u,x) \in E\\
    0 & \text{else}
    \end{cases}
\end{equation}
where $U(\cdot,\cdot)$ is the unnormalized transition probability, and $Z$ is the normalizing constant. The sampling strategy $O$ could be, for example, Breadth-First Sampling (BFS), Depth-First Sampling (DFS), etc., generating different transition probabilities. 
We follow \textit{node2vec} to use a biased $2^{nd}$ order random walk as the sampling strategy. In our unweighted road network graph,
\begin{equation}
    U(u,x)=\alpha_{pq}(u,x)=\begin{cases}
    \frac{1}{p} &\text{if } d_{wx}=0\\
    1 &\text{if } d_{wx}=1\\
    \frac{1}{q} &\text{if } d_{wx}=2
    \end{cases}
\end{equation}
where $p$, $q$ are two parameters of the random walk which control how fast the walk explores, $w$ is the predecessor of $u$, and $d_{wx}$ is the distance between nodes $w$ and $x$. 

After obtaining the mapping function $g$, we use it to embed nodes when extracting features from a state. For action representation, we use a learnable embedding layer. We apply this setting because it can keep flexibility in learning while leveraging graph information.
\begin{figure}
    \centering
    \includegraphics[width=0.48\textwidth]{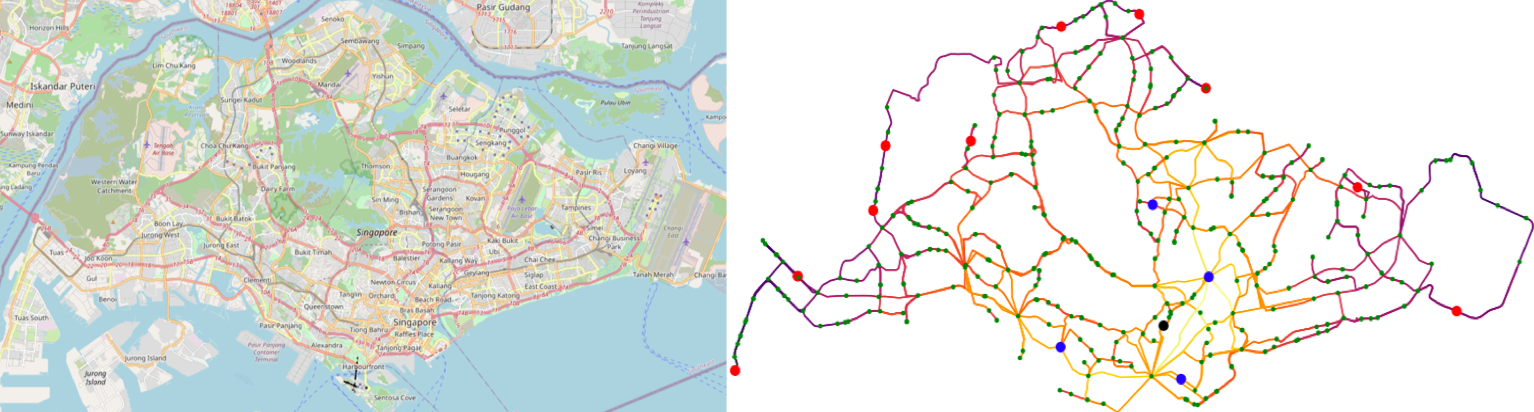}
    \caption{Singapore map and the extracted road network.}
    \label{sg_map}
\end{figure}

\subsection{Discussion about Other Applicable Games}
The proposed method is suitable for solving games whose legal actions vary significantly with states. Since the legal actions at different states can be very different, the overall action spaces of these games are usually enormous, leading to inefficiency of existing approaches. Our method is a mitigation for the problem. Apart from NSGs, our method can be applied to games that involve high-dimensional control. In such games, the overall legal action space is a combination of each dimension's legal action space, changes of legal actions in each dimension will accumulate and may lead to a significant change in the overall legal action space. 
Team-Goofspiel (Section \ref{adapt}) is an example for that. 
Another potential application scenario of our method is text-based games whose actions are defined based on natural languages, i.e., the action spaces are composed of sequences of words from a fixed size and large dictionary.  There are constraints on actions at each state to generate a meaningful command (sequence of words), e.g., “open door”, “turn left”. For different states, e.g., "open" or "turn", the corresponding legal action spaces are very different (“door, box, book, $\dots$” or “left, right, $\dots$”). Our method can 
be applied to them by firstly learning states and legal action representations and then mapping state-action pair representations to values which represent Q-values or probabilities. High-level actions can be defined, for example, as phrases or sentences, and learning embeddings for each word is also reasonable. 

\section{Experimental Evaluation}
We firstly evaluate our algorithm on large-scale NSGs. Then, we perform ablation studies to understand how each component of NSG-NFSP affects the results. Finally, we justify the adaptability of our method to games in other domains. Experiments are performed on a server with a 10-core 3.3GHz Intel i9-9820X CPU and an NVIDIA RTX 2080 Ti GPU.

\subsection{Large-Scale NSGs}
\label{largeNSG}
We evaluate our algorithm in NSGs played on both artificially generated networks and real-world road networks.

\paragraph{Artificially Generated Networks}
 We generate the evaluation network by the grid model with random edges \cite{randomgrid}. Concretely, we sample a $15\times15$ grid whose horizontal/vertical edges appear with probability 0.4 and diagonal edges appear with probability 0.1. We set the initial location of the attacker at the center of the grid, and let the defender controls 4 security resources which are distributed uniformly on the network at the beginning\footnote{NSG-NFSP allows the players to do stochastic initialization, by taking the initialization as the first-step action.}.
 There are 10 target nodes located randomly at the border. We set the time horizon $T$ as 70, 90, and 300, to create NSGs with different scales.

 We compare our algorithm with two heuristic defender policies, namely uniform policy and greedy policy, as well as a state-of-the-art algorithm, IGRS++ \cite{yz19}. For the uniform policy, the defender assigns equal probability to each legal action at a state. For the greedy policy, all of the defender's security resources always move along the shortest path to the attacker's current location. When evaluating the performance, because the game sizes are very large, it is intractable to calculate exact worst-case defender utility. We overcome this by using approximate worst-case defender utility. Specifically, we use a DQN attacker to best respond to the defender and calculate defender utility under this scenario to approximate the worst-case defender utility. We train the DQN attacker for $2\times10^5$ episodes, store the best model, and load the best model to play with the defender for another 2000 episodes to obtain the final results.
 We run the NSG-NFSP for $5\times10^6$ episodes, which takes around 3-4 days. Despite that the method consumes non-trivial resources when training, it can realize real-time inference, making its deployment possible. Neural network structures and hyper-parameters are included in the appendix.
 As in Table \ref{largegrid}, our method, NSG-NFSP, outperforms the baselines in all three large-scale NSGs. The state-of-the-art algorithm, IGRS++, cannot execute. The reason is that, as an incremental strategy generation algorithm, IGRS++ requires the attacking paths to be enumerable. However, for all of the three settings, it is impossible to enumerate attacking paths. 
 We try to enumerate attacking paths for the case $T=70$ on a machine with 32G RAM, but after all the memory is occupied, the enumeration does not end.
 Applying counterfactual regret minimization (CFR)~\cite{cfr} or its variants \cite{cfrvariance,deepcfr} to solve large-scale NSGs is also infeasible, and we discuss about it in the appendix.
 Note that the approximate worst-case defender utilities for the greedy policy are always 0, which means that the DQN attacker can always find at least one path to a target node such that the defender with greedy policy cannot prevent him. 

\begin{table}
\begin{subtable}{1\linewidth}
\centering
\scalebox{0.79}{
\begin{tabular}{|l|c|c|c|c|}
\hline 
 & \multirow{1}{*}{\begin{tabular}[c]{@{}c@{}}NSG-NFSP \end{tabular}} & \multirow{1}{*}{\begin{tabular}[c]{@{}c@{}}Uniform 
 policy\end{tabular}} & \multirow{1}{*}{\begin{tabular}[c]{@{}c@{}}Greedy  Policy\end{tabular}}&\multirow{1}{*}{\begin{tabular}[c]{@{}c@{}}IGRS++ \end{tabular}}\\ 
   \hline\hline
T=70  & \textbf{0.1770 $\pm$ 0.0168}   & 0.0730 $\pm$   0.0114      & 0 $\pm$ 0    &   OOM   \\ \hline
T=90  & \textbf{0.1205 $\pm$ 0.0143} &        0.0715  $\pm$ 0.0111     & 0 $\pm$   0  &    OOM   \\ \hline
T=300 & \textbf{0.0825 $\pm$ 0.0120}   &        0.0675 $\pm$ 0.0110       & 0  $\pm$ 0   &     OOM   \\ \hline 
\end{tabular}
}
\caption{Synthetic network}
\vspace{3mm}
\label{largegrid}
\centering
\scalebox{0.79}{
\begin{tabular}{|l|c|c|c|c|}
\hline 
 & \multirow{1}{*}{\begin{tabular}[c]{@{}c@{}}NSG-NFSP \end{tabular}} & \multirow{1}{*}{\begin{tabular}[c]{@{}c@{}}Uniform 
 Policy\end{tabular}} & \multirow{1}{*}{\begin{tabular}[c]{@{}c@{}}Greedy  Policy\end{tabular}}&\multirow{1}{*}{\begin{tabular}[c]{@{}c@{}}IGRS++ \end{tabular}}\\ 
   \hline\hline
T=30  & \textbf{0.1225 $\pm$ 0.0140}   & 0.0230 $\pm$  0.0066        & 0   $\pm$ 0   &     OOM   \\ \hline
T=300 & \textbf{0.0720 $\pm$ 0.0113}  &      0.0055  $\pm$ 0.0032     & 0 $\pm$ 0  &     OOM      \\ \hline 
\end{tabular}
}
\caption{Real-world road network}
\label{largesg}
\end{subtable}%
\caption{Approximate worst-case defender utilities in NSGs with different time horizons. The ``$\pm$'' indicates $95\%$
    confidence intervals over the 2000 testing episodes. OOM stands for Out of Memory.}
\end{table}

\paragraph{Real-World Road Networks}
As in Figure \ref{sg_map}, we extract highways, primary roads and the corresponding intersections from Singapore map via OSMnx \cite{osmnx}. There are 372 nodes and 1470 edges.
Edges are colored according to closeness centrality. The brighter the color, the closer the edges are to the center. The initial position of the attacker, marked in the dark point, locates near the center. There are 4 security resources, marked in blue points. Those target nodes, denoting exits of the map, are marked in red points. We test out that $T=30$ will lead to attacking/escaping paths unenumerable, and we set the time horizon at 30 and 300.
The NFSP defender is trained for $5\times10^6$ episodes which takes around a week to finish. We keep the evaluation settings the same as in synthetic networks for $T=30$ because we find the settings work well. For $T=300$, where training a DQN attacker is more difficult, we fine-tune hyperparameters to enhance the DQN attacker's performance.
As presented in Table \ref{largesg}, our method significantly outperforms the baselines in both settings. 
Additional experiments on the Manhattan map also show that our method outperforms the baselines.
More experiment results are in the appendix.
\subsection{Ablation Studies}
\label{as}
We perform ablation studies on $7\times7$ and $15\times15$ randomly generated grids, with $T$ at 7 and 15, respectively.

\begin{figure}
     \begin{subfigure}[t]{0.22\textwidth}
         \centering
         \includegraphics[width=1\textwidth]{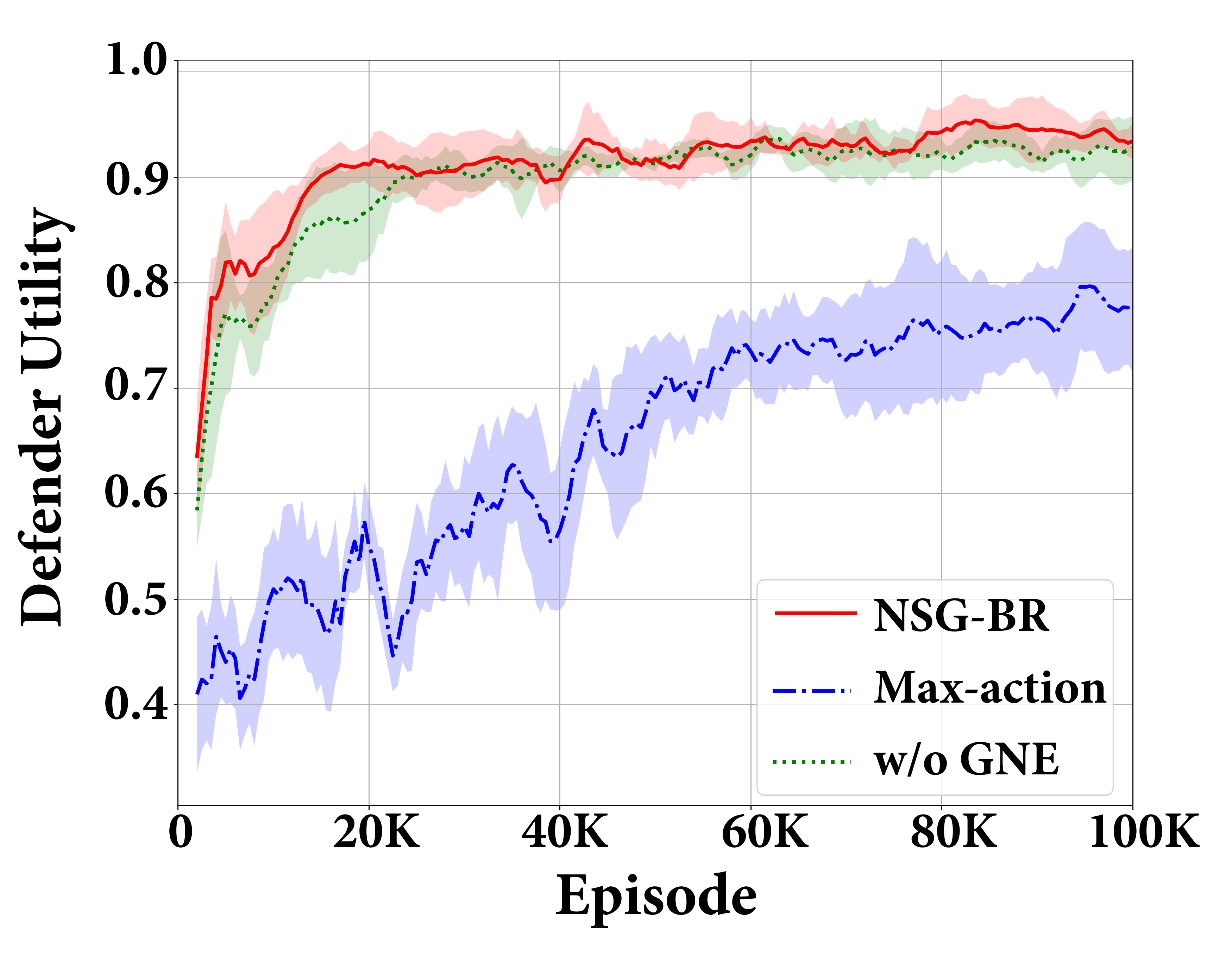}
         \caption{$7\times7$ grid (T=7)}
         \label{br_grid7}
     \end{subfigure}
     \hfill
     \begin{subfigure}[t]{0.22\textwidth}
         \centering
         \includegraphics[width=1\textwidth]{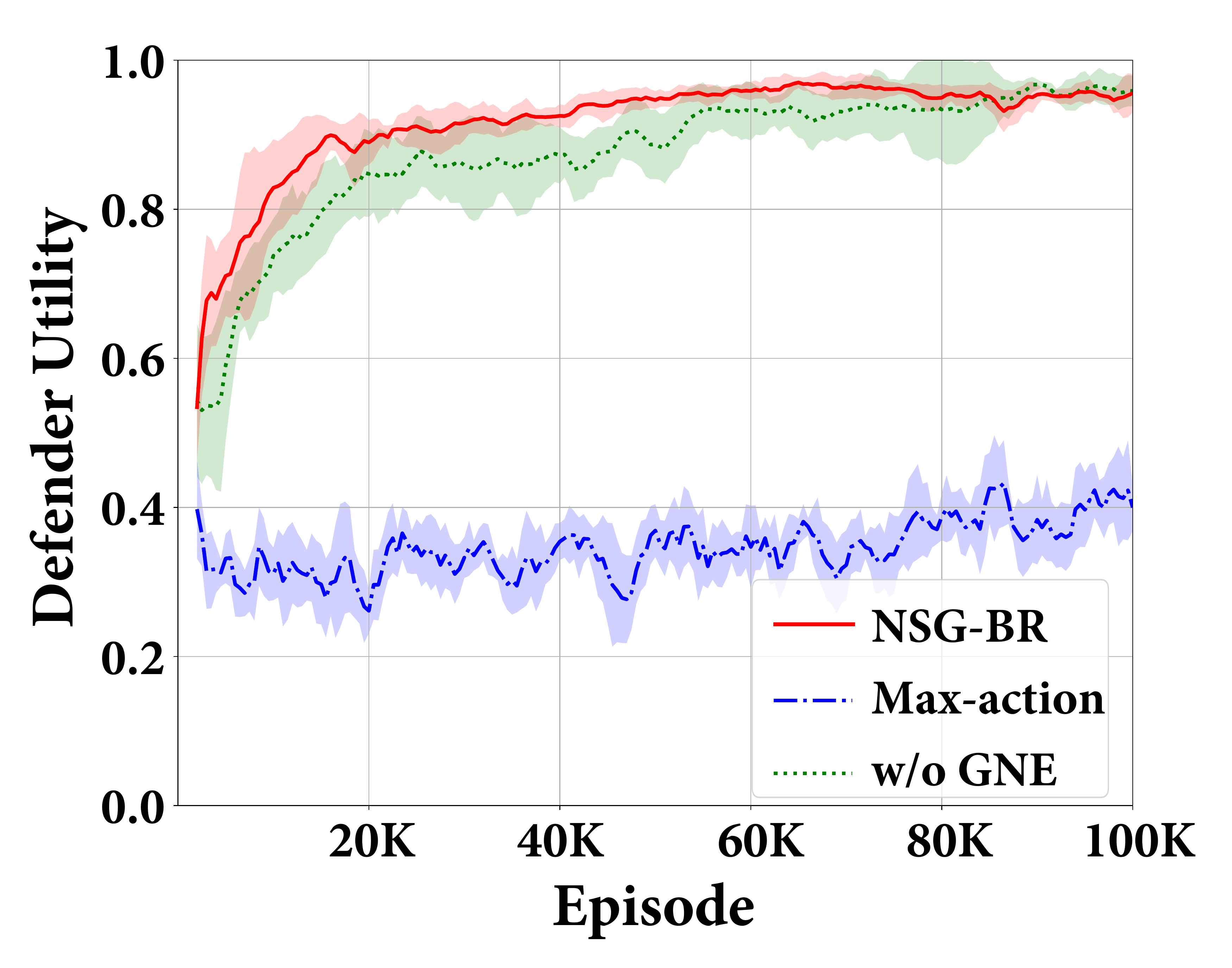}
         \caption{$15\times15$ grid (T=15)}
         \label{br_grid15}
     \end{subfigure}
     \caption{The learning curves of the defender against an attacker with uniform policy on synthetic networks,
      averaged across 5 runs.}
      \label{br}
\end{figure}

\paragraph{Best Response Approximation} We try to evaluate whether the proposed network architecture, NSG-BR, is able to enhance best response approximation. We set the policy of the attacker to be uniform, and let the defender best respond to him. 
We compare NSG-BR with a naive implementation, max-action DQN, which fixes the output dimension of BR policy network equal to the maximum number of legal actions. 
As in Figure \ref{br}, the performance of max-action DQN is significantly worse than NSG-BR in both settings.
We further explore the effect of pre-defined graph node embedding (GNE). We replace GNE with a learnable embedding layer (w/o GNE).
Results show that, in simple graph (the $7\times7$ grid), the learning curves have no obvious difference. However, in complex graph (the $15\times15$ grid), GNE does benefit the training. If created properly, GNE can not only speed up the training but also make the process more stable. 

\begin{figure}
     \begin{subfigure}[t]{0.235\textwidth}
         \centering
         \includegraphics[width=1\textwidth]{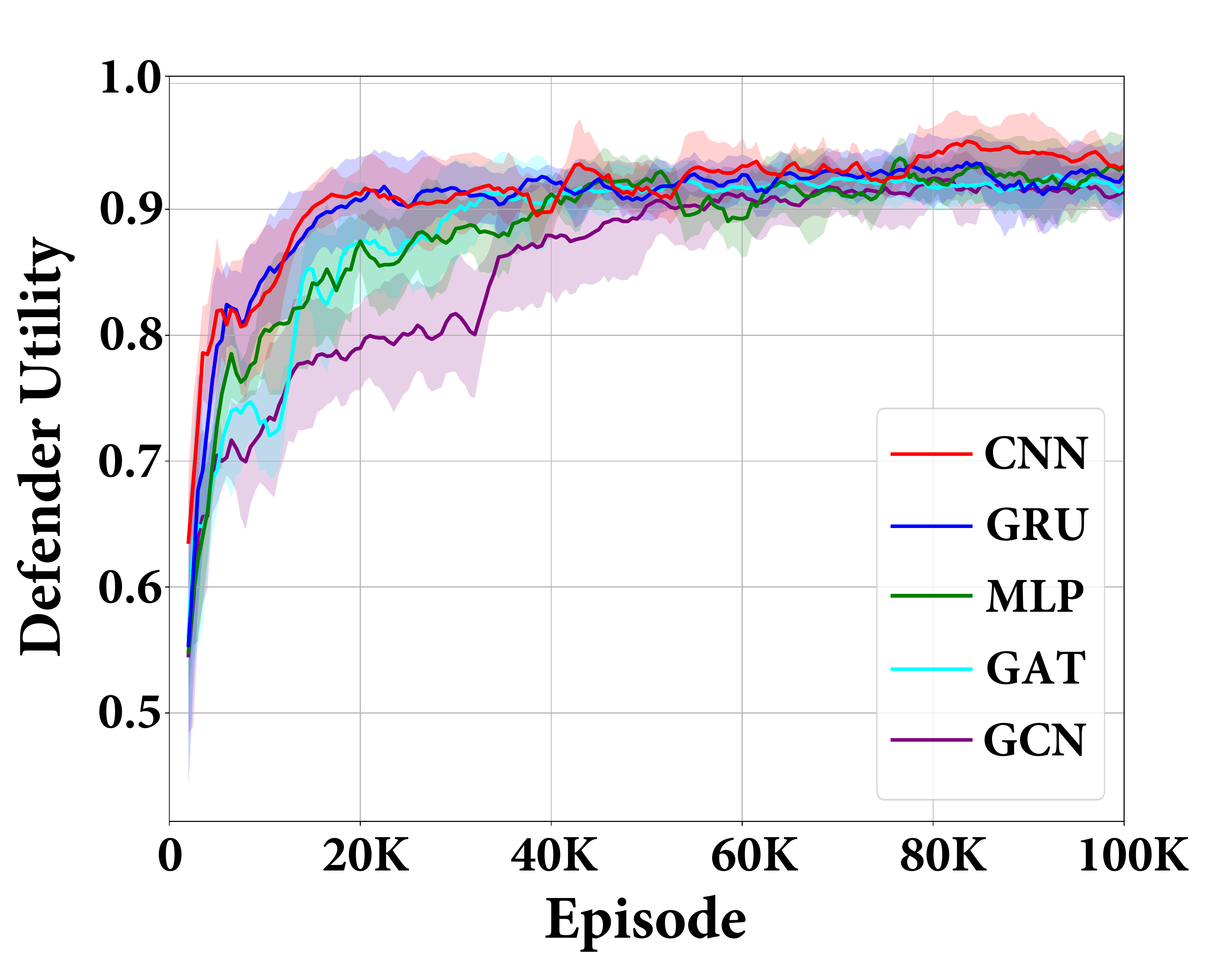}
         \caption{The learning curves}
         \label{lc}
     \end{subfigure}
     \hfill
     \begin{subfigure}[t]{0.235\textwidth}
         \centering
         \includegraphics[width=1\textwidth]{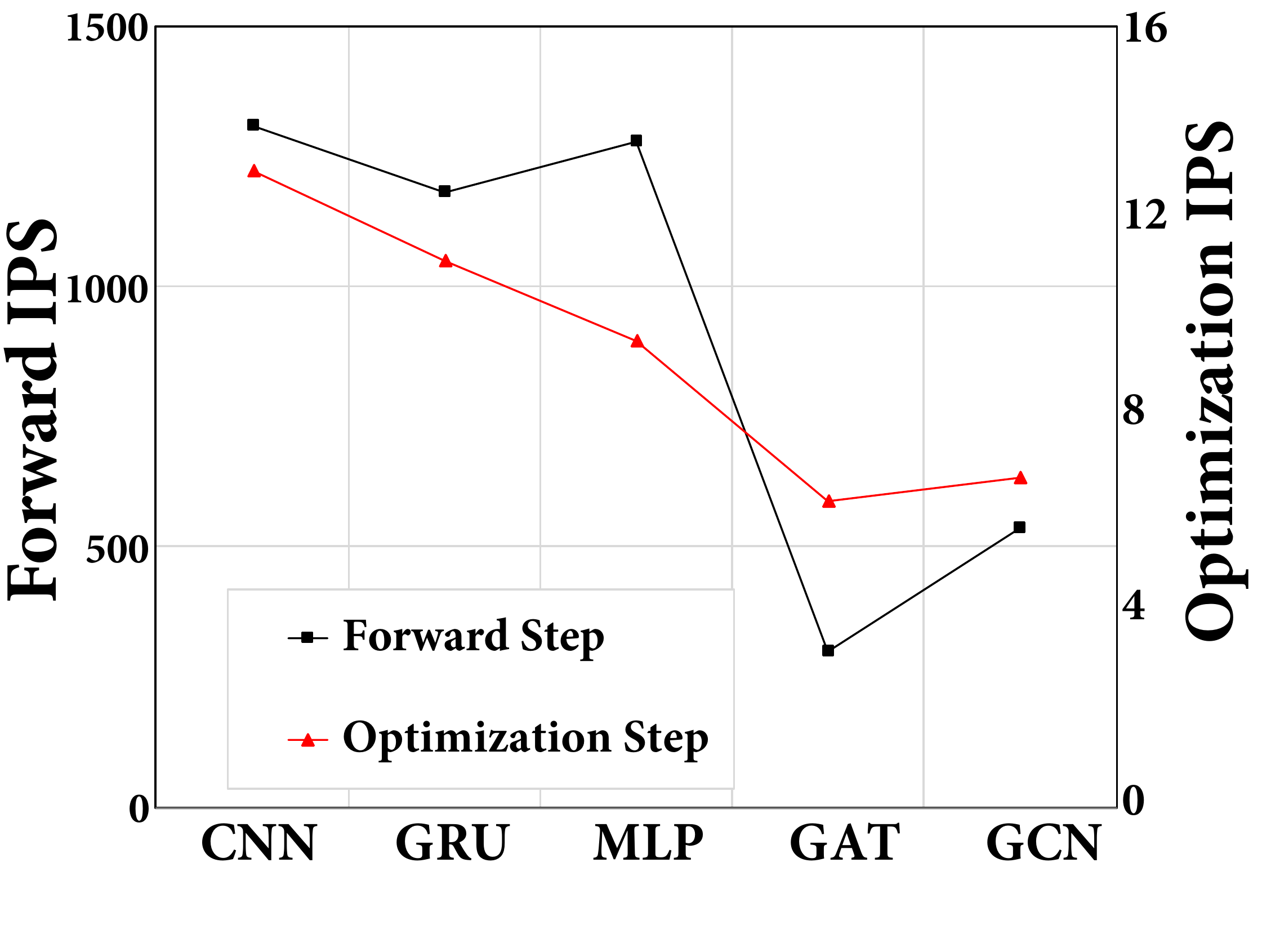}
         \caption{Iterations per second}
         \label{ips}
     \end{subfigure}
     \caption{Performance of sequence encoders on the $7\times7$ grid.
     }
      \label{seq}
\end{figure}

\paragraph{Comparisons of Different Sequence Encoders}
We compare the performance of different types of sequence encoders for extracting state features. The encoders include: i) GatedCNN (CNN), a CNN-based approach (what we use in NSG-NFSP); ii) Gated Recurrent Unit (GRU) \cite{gru}; iii) Graph Convolution Network (GCN)~\cite{gcn}; iv) Graph Attention Network (GAT)~\cite{gat}; and v) Multi-Layer Perceptron (MLP). 
We evaluate the average defender utility on the synthetic $7\times7$ grid, setting the policy of the attacker to be uniform. 
The learning curves of the defender are presented in Figure \ref{lc}. We can find that GatedCNN obtains superior performance over other sequence encoders. 
We further test the running speed by measuring number of iterations per second (IPS) for both the forward process and the optimization process (batchsize=256).
The forward process affects the speed to collect experiences, and the optimization process affects the speed to optimize parameters of DNNs. As in Figure \ref{ips}, GatedCNN is the fastest of all the sequence encoders.
 
\begin{figure}
     \begin{subfigure}[t]{0.23\textwidth}
         \centering
         \includegraphics[width=1\textwidth]{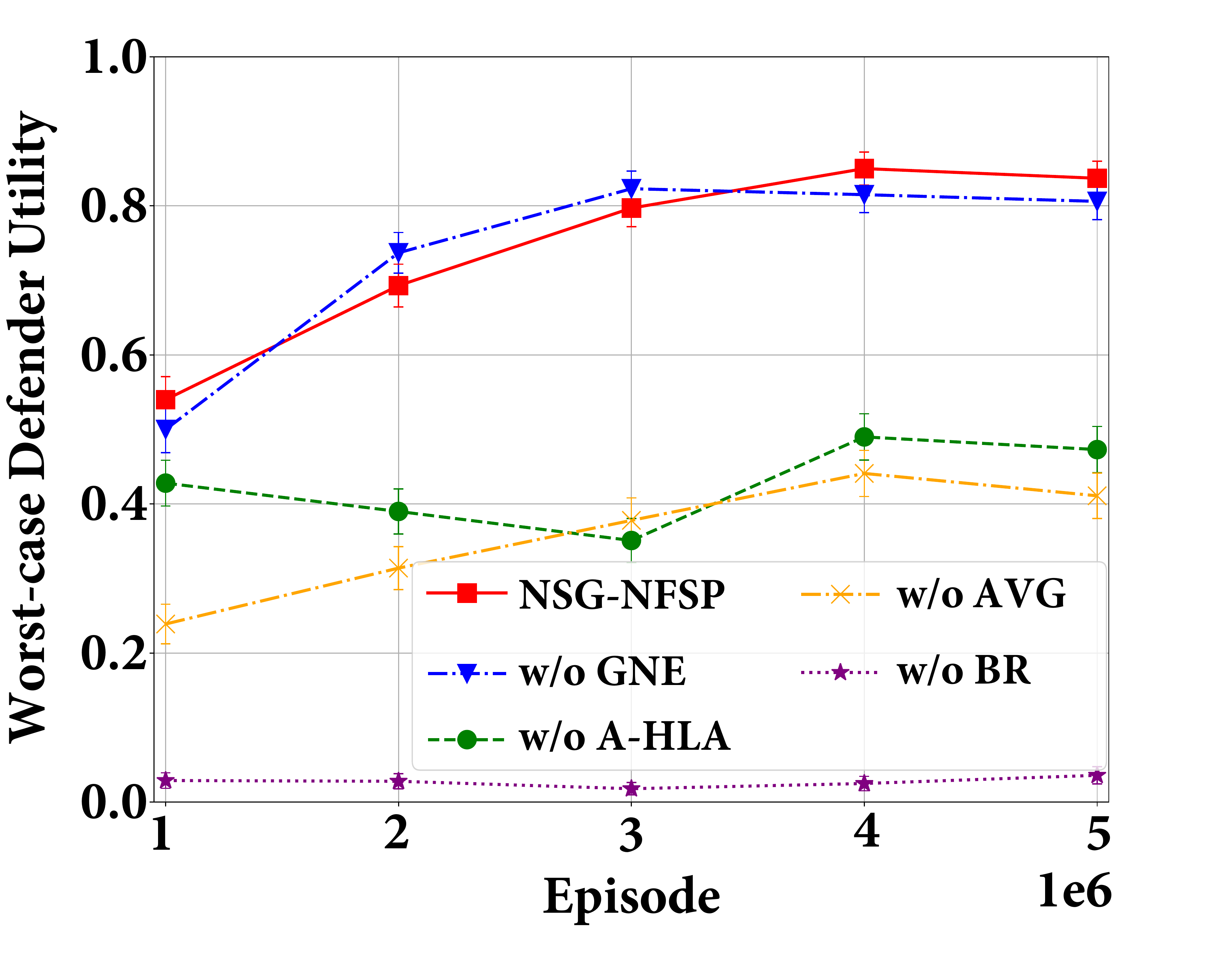}
         \caption{$7\times7$ grid (T=7)}
         \label{uti_grid7}
     \end{subfigure}
     \hfill
     \begin{subfigure}[t]{0.23\textwidth}
         \centering
         \includegraphics[width=1\textwidth]{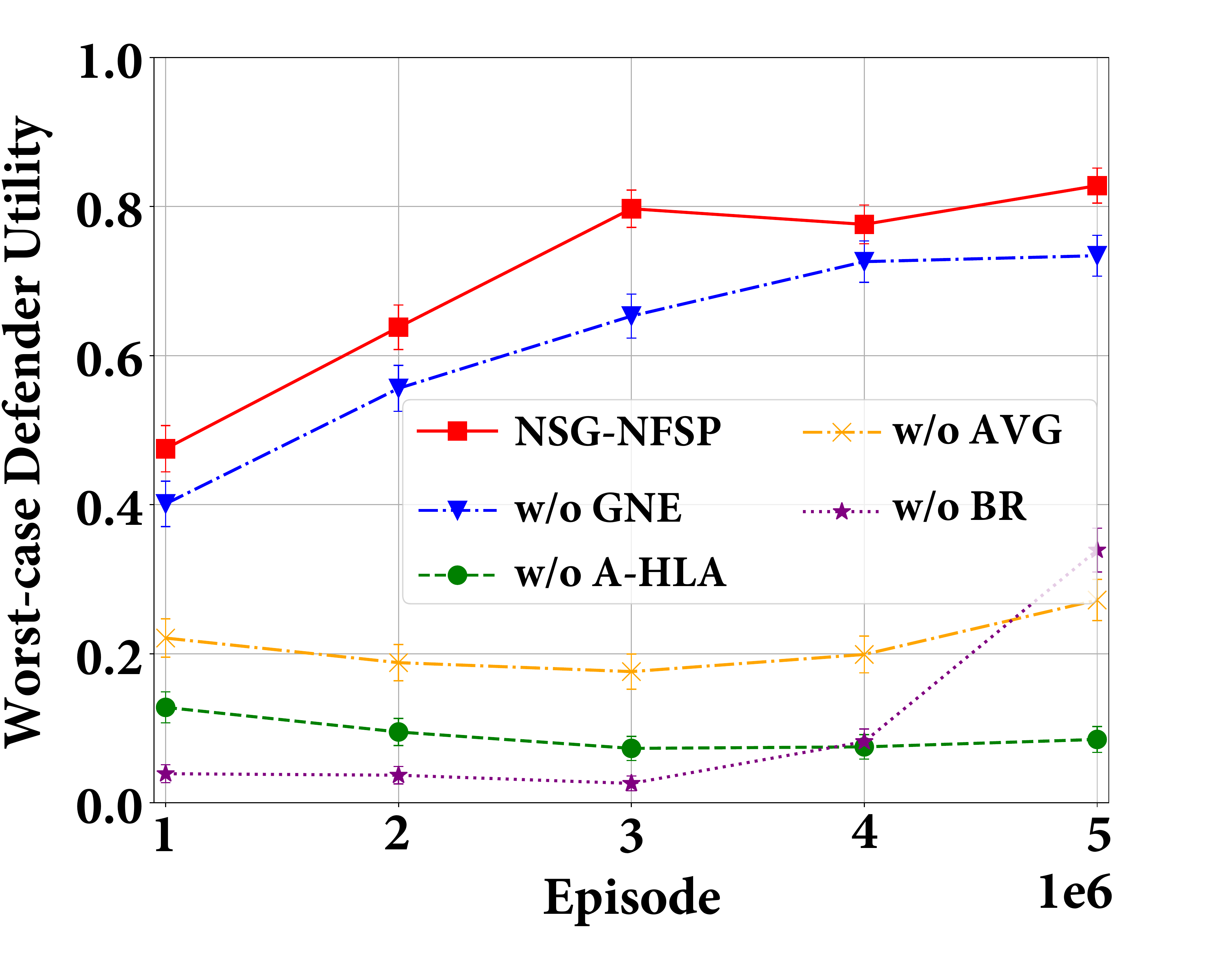}
         \caption{$15\times15$ grid (T=15)}
         \label{uti_grid15}
     \end{subfigure}
     \caption{Ablation studies regarding each component of NSG-NFSP.
     }
      \label{uti}
\end{figure}

\paragraph{Worst-Case Defender Utility} To show how each component of NSG-NFSP affects the performance, we replace i) best response policy approximation module (Section \ref{abrp}) with max-action DQN (w/o BR); ii) average policy approximation module (Section \ref{aap}) with max-action network (w/o AVG); iii) the  attacker in high-level control (Section \ref{mtap}) with the original low-level implementation (w/o A-HLA); and iv) graph node embeddings (Section \ref{egne}) with a learnable embedding layer (w/o GNE). As in Figure \ref{uti}, replacing any of BR, AVG and A-HLA will cause a dramatic drop in performance. As for GNE, it shows a slight performance improvement in simple networks (the $7\times7$ grid), but in complex networks (the $15\times15$ grids), the enhancement is significant.  NSG-NFSP achieves a performance of around 0.8 for the both games. Since the worst-case defender utility is upper-bounded by 1 (the defender wins the game definitely), the results demonstrate near-optimal solution quality of our method.

\subsection{Adaptability}
\label{adapt}
\begin{figure}
     \begin{subfigure}[t]{0.235\textwidth}
         \centering
         \includegraphics[width=1\textwidth]{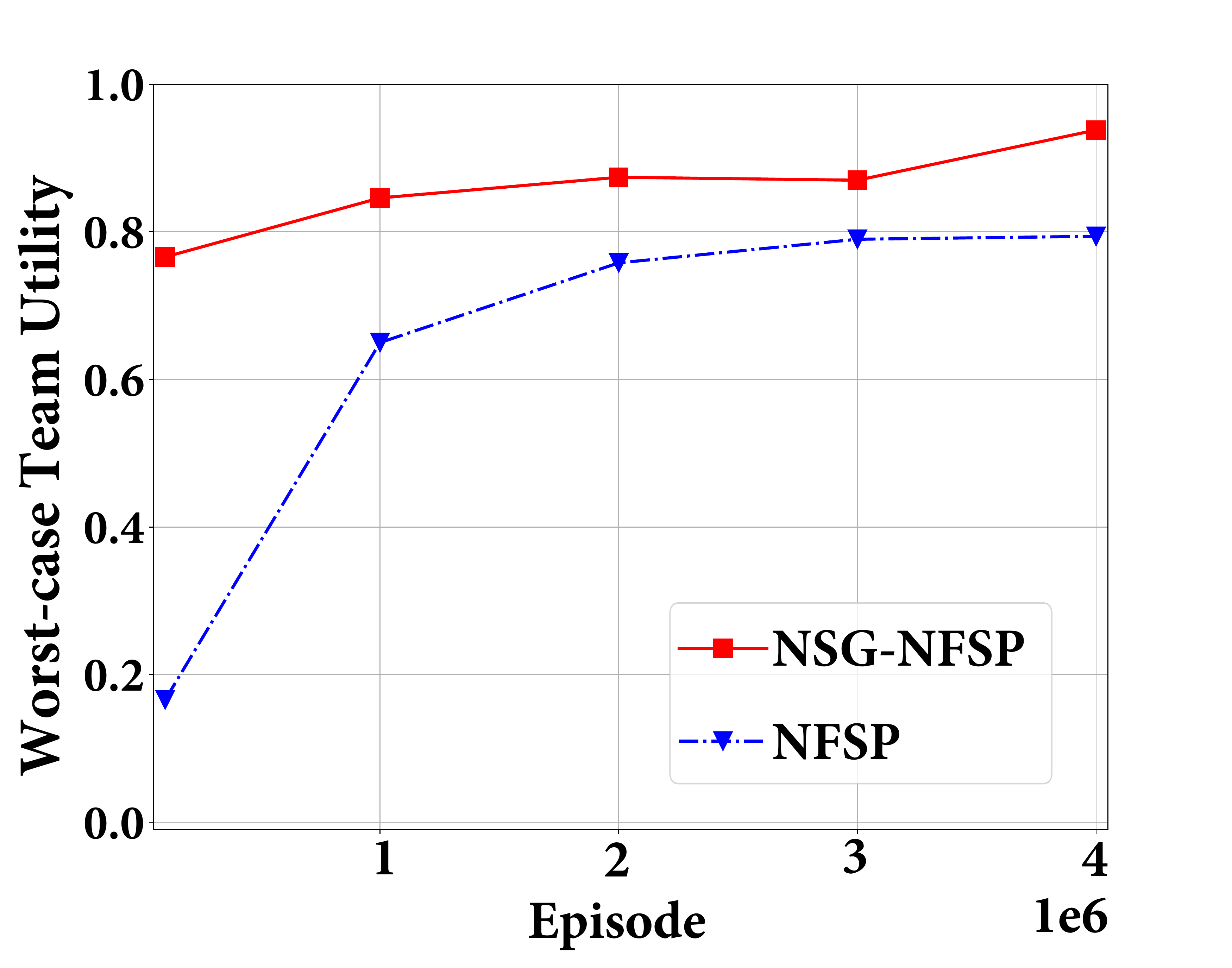}
         \caption{MAX mode}
         \label{goof_max}
     \end{subfigure}
     \hfill
     \begin{subfigure}[t]{0.235\textwidth}
         \centering
         \includegraphics[width=1\textwidth]{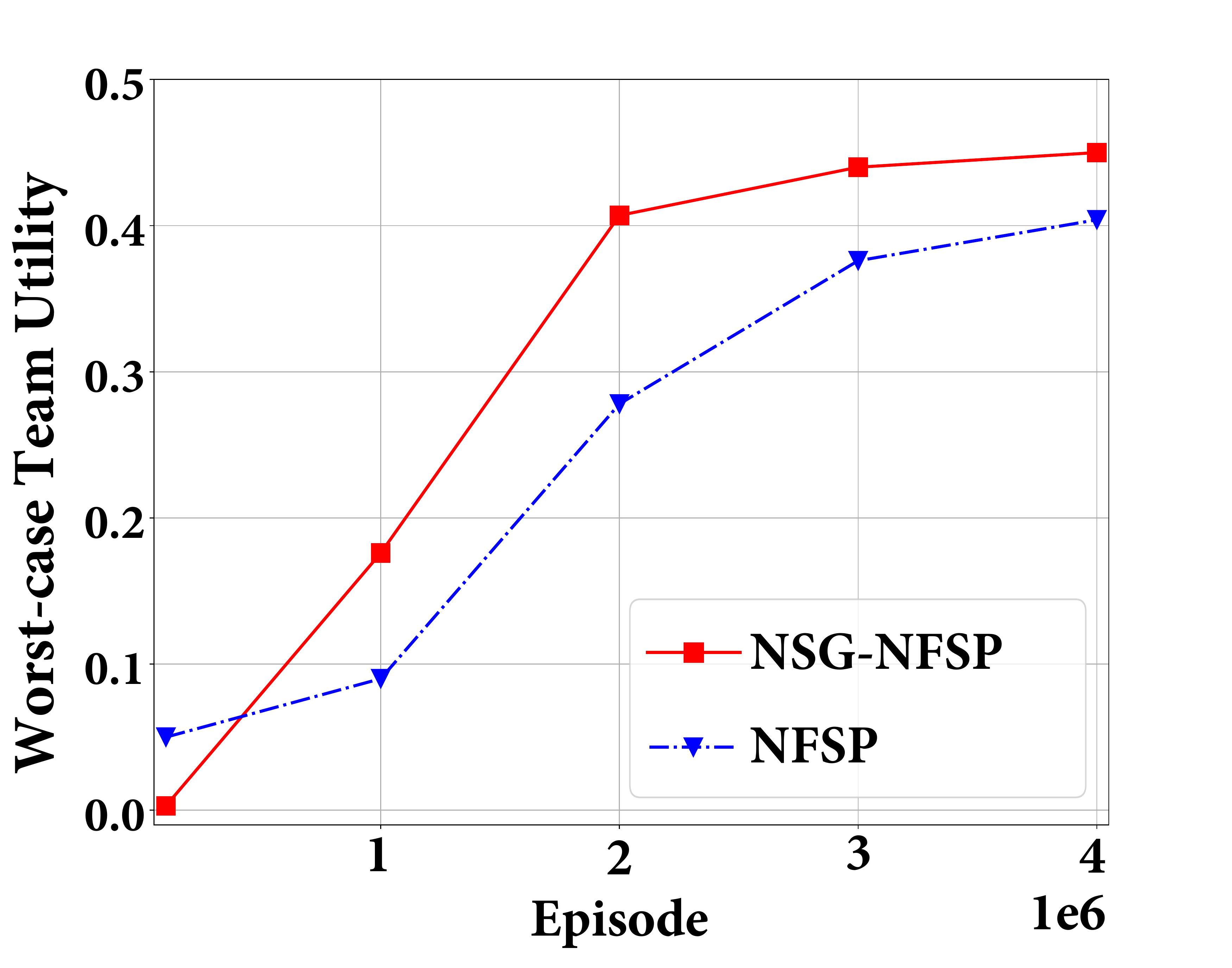}
         \caption{AVERAGE mode}
         \label{goof_mean}
     \end{subfigure}
     \caption{The worst-case team utility in 4-rank 2-member Team-Goofspiel under MAX and AVERAGE game modes.
     }
      \label{goof}
\end{figure}

Our method is suitable for games whose legal action spaces or their sizes vary significantly with states, including but not limited to NSGs. To justify this, we conduct experiments on an extended version of Goofspiel \cite{goofspiel}, a popular poker game. Specifically, we modify Goofspiel to be played between a team with several members and a single player, as opposed to two single players. We name the modified game as Team-Goofspiel. In $k$-rank $n$-member Team-Goofspiel, there are $k$ rounds and the team consists of $n$ members. At each round $t$ ($t=1,\dots,k$), $n$ team members and the single player place bids for a prize of value $t$. The possible bids are $1,\dots,k$ and each bid can be placed exactly once (the action space of the team player decays rapidly with respect to $t$, containing $(k+1-t)^n$ legal actions in total). The player with higher bid wins the prize of the current round; if the bids are equal, no player wins the prize. Both players can only observe the outcome of each round but not the bids. The single player will win the game if its total prize is greater than the team's prize at the end of the game, otherwise the team wins. The winner will obtain a utility of 1. We design two game modes for Team-Goofspiel, namely the MAX mode and the AVERAGE mode, where the bid of the team is determined by taking the maximum or the average of all team members' bids, respectively. We set $k=4$ and $n=2$ in the experiments. As in Figure \ref{goof}, our method obtains superior performance compared to the vanilla NFSP algorithm in both the MAX mode and the AVERAGE mode, demonstrating its good adaptability\footnote{
Despite that it is easy to define high-level actions, e.g., bids sequences, and learning bids embeddings to reflect their numerical relationships is reasonable, we omit these in our implementation because they are domain-specific.}.
\section{Conclusions}
In this paper, we propose a novel learning paradigm, NSG-NFSP, for finding NE in large-scale extensive-form NSGs. The algorithm trains DNNs to map state-action pairs to values, which may represent Q-values or probabilities. It enhances the performance by enabling the NFSP attacker with high-level actions and learning efficient graph node embeddings.
Our method significantly outperforms state-of-the-art algorithms in both scalability and solution quality. 

\section*{Acknowledgements}
This research is partially supported by Singtel Cognitive and Artificial Intelligence Lab for Enterprises (SCALE@NTU), which is a collaboration between Singapore Telecommunications Limited (Singtel) and Nanyang Technological University (NTU) that is funded by the Singapore Government through the Industry Alignment Fund – Industry Collaboration Projects Grant.

\bibliographystyle{named}
\bibliography{ijcai21}

\begin{appendices}
\onecolumn

\section{Further Discussion about Related Works}
\label{further_rw}
The main challenges of applying NFSP to NSGs lie in two aspects: i) approximating best response policy and ii) approximating average policy when action space is large and lots of illegal actions exist at each state. We introduce the works which are relevant to our solutions of these two problems separately. 

\paragraph{Deep RL in Large Discrete Action Spaces}
Deep RL formulates the computation of best response as a Markov Decision Process (MDP). An MDP is defined by a 5-tuple $\langle \mathcal{S},\mathcal{A},r, P ,\gamma \rangle$, where $\mathcal{S}$ is a set of states, $\mathcal{A}$ is a set of actions, $r(s,a)$ is a reward function, 
$P(s'|s,a)$ is a transition function, and $\gamma$ is a discount factor. The goal of learning is to find the policy $\pi$ mapping states to actions that maximizes the expected discounted total reward. The action-state value function $Q^{\pi}(s,a)$ is an estimate of the expected future reward that can be obtained when starting from $s$, acting $a$, and obeying policy $\pi$ (in following steps).
The optimal $Q^{*}(s,a)$ is determined by solving the Bellman equation:
\begin{equation}
    Q^{*}(s,a)=\mathbb{E}\Big[r(s,a)+\gamma\sum_{s'}P(s'|s,a)\max_{a'\in\mathcal{A}(s')}Q^{*}(s',a')\Big]
\end{equation}
The optimal policy is $\hat{\pi}(s)=\arg\max_{a\in\mathcal{A}(s)}Q^*(s,a)$. 

DQN \cite{dqn} proposes to use a DNN to approximate $Q(s,a)$. It outputs Q-values for all actions, thus unable to work in large action spaces. To allow deep RL in large discrete action spaces, Wolpertinger Architecture (WA) \cite{la15} proposes to transform discrete actions to a continuous space by leveraging prior information and predict optimal action by applying a nearest-neighbor search. 
RA \cite{ar} extends WA by learning action embeddings and a transformation function to transform action embeddings back to actual actions. Though these two methods allow generalization over similar actions, in NSGs where there are plenty of illegal actions at each state, it is hard to pre-define or train an action transformation function which is able to precisely hit legal actions. 
DRRN \cite{acl} is another approach, which learns representations for states and actions respectively and models Q-values as an inner product of state-action representations pairs. Act2Vec \cite{ar} extends DRRN by using data from expert demonstrations to learn action representations. Our method in learning best response policy is inspired by DRRN-based approaches. This kind of approaches assume that actions are in natural languages domain. However, in NSGs, actions are defined based on graph nodes. Thus, we should adapt the approaches to make them suitable for NSGs.

\paragraph{Metric-Based Few-Shot Learning}
Our method in approximating average policy is inspired by Few-Shot Learning (FSL). 
It is proposed to mitigate the problem that learning a universal classifier over all existent categories is impossible.
The scenario of FSL is somewhat similar to the problem we meet in learning average policy: when facing an enormous action space and there are lots of illegal actions at each state, it is desirable to avoid assigning a distribution to all actions, and generating classifier over legal actions set $\mathcal{A}(s) \subseteq \mathcal{A}$ for each state is preferred. 

Metric-based FSL is the most relevant category to our approach. They transform samples to a space where samples can be classified by comparing some metrics. Effective metrics could be cosine similarity, Euclidean distance, etc.
Relation Network \cite{relationnet} enables the metric to be learnable. When learning average policy, a learnable metric is preferred because a policy can be very complex.
Thus, we adopt a network structure similar to Relation Network when approximating average policy. 
Note that we do not require the network to generalize to new classes, i.e., non-existent actions, though it is the main purpose of few-shot learning.

\section{The Overall Algorithm}
\label{overallalgo}
We present the overall algorithm in Algorithm \ref{algo}. 
In line 1, we calculate node embeddings by applying $node2vec$ (introduced in Section \ref{egne}). From lines 2 to 3, we do initialization for the defender and the attacker. In line 5, both the defender and the attacker sample to determine the behaviour for the current episode. From lines 6 to 8, the attacker samples to decide whether to explore in the current episode or not. From lines 9 to 10, the attacker determines the escaping path for the current episode. From lines 11 to 19, the defender collects transitions for training its policy networks. From lines 20 to 23, the attacker collects data for learning its policy.  From lines 24 to 26, the defender updates its policy networks (introduced in Section \ref{abrp} and Section \ref{aap}). From lines 27 to 28, the attacker updates its policy (introduced in Section \ref{mtap}).
The defender's average policy network is the output of the algorithm.
\begin{table*}[ht]
\centering
\begin{tabular}{l|c|c|c|c|c|c|c|c|c}
\hline
Number of Nodes Touched     & 1e5  & 2e5   & 3e5   & 4e5   & 5e5            & 6e5   & 7e5   & 8e5   & 9e5   \\ \hline
Worst-case Defender Utility & 0.210 & 0.213 & 0.184 & 0.208 & \textbf{0.223} & 0.197 & 0.194 & 0.174 & 0.174 \\ \hline
\end{tabular}
\caption{The worst-case defender utilities for OS-CFR on the synthetic $7\times7$ grid ($T=7$).}
\label{oscfr}
\end{table*}

\section{Neural Network Structures and Hyperparameters}
\label{nnstructure}
Graph nodes are embedded into vectors of length 32. For the state representation network, we use a two-layer MLP with 64 hidden units and 64 outputs to extract features for resource locations. The GatedCNN module is composed of two identical 1D-convolutional layers with 64 kernels of size 3 and stride 1. One of the convolutional layers is activated by sigmoid function, and the result is point-wisely multiplied with the output of the other convolutional layer. The GatedCNN module is followed by a global max-pooling layer. The action representation network is a two-layer MLP with 64 hidden units and 64 outputs. State and action features are concatenated and sent to a two-layer MLP with 64 hidden units to predict Q-values or probabilities. Unless otherwise specified, all layers are separated with ReLU nonlinearities.

We use the Adam optimizer with a learning rate of $1\times10^{-4}$ to optimize the average policy network. The RMSprop optimizer with the same learning rate is used for optimizing the BR policy network. Gradients are clipped according to the 2-norm to a threshold of 1. We update the BR policy network every 4 episodes with batch size at 128 and update the average policy network every 32 episodes with batch size at 256. The capacities of $\mathcal{M}_{RL}$ and $\mathcal{M}_{SL}$ are $5\times10^5$ and $1\times10^7$, respectively. The parameters of the target network are copied from the BR policy network every  $1\times10^3$ episodes. The attacker updates the MAB at the end of each episode, 
and the MAB keeps a record of the latest $1\times10^4$ plays. The AVGer is updated every $1\times10^3$ episodes. We set the anticipatory parameter $\eta$ as 0.1 and the attacker exploration probability as 0.1. 

\section{Discussion about CFR-Based Approaches}
\label{appcfr}
NSGs are extensive-form games (EFGs). Therefore, it is natural to think of applying Counterfactual Regret Minimization (CFR), which is one of the leading technology for EFGs, to solve NSGs. However, vanilla CFR requires traversing the whole game trees which is infeasible in large-scale NSGs whose game trees are prohibitively deep and wide. We test a sampling-based derivative of CFR, Outcome-sampling CFR (OS-CFR) with variance reduction \cite{cfrvariance}, on the synthetic $7\times7$ grid (in Section \ref{as}). Results are presented in Table \ref{oscfr}. OS-CFR uses up all memories after it touches around 900,000 information nodes (the memory limitation is the main issue for tabular approaches). The best performance for OS-CFR is 0.223. For NSG-NFSP, as presented in Figure \ref{uti_grid7}, the best performance is 0.85, which is significantly better than OS-CFR. We also try DeepCFR \cite{deepcfr}, a deep learning version of CFR, but it fails to demonstrate obvious learning effects in NSGs. We think the reason is that the heavy variances of estimated regret values cause difficulties in learning.
\begin{table*}
\centering
\begin{tabular}{l|ccccc}
\hline
 Episodes     & 1e6 & 2e6 & 3e6 & 4e6 & 5e6 \\ \hline\hline
T=70  & \textbf{0.1770 $\pm$ 0.0168} & 0.1285 $\pm$ 0.0147          & 0.0630 $\pm$ 0.0107          & 0.0905 $\pm$ 0.0126  & 0.0570 $\pm$ 0.0102   \\ \hline
T=90  & 0.0730 $\pm$ 0.0114          & 0.1070 $\pm$ 0.0136           & \textbf{0.1205 $\pm$ 0.0143} & 0.0955 $\pm$ 0.0129   & 0.0660 $\pm$ 0.0109   \\ \hline
T=300 & \textbf{0.0825 $\pm$ 0.0120}          & 0.0615 $\pm$ 0.0105 & 0.0545 $\pm$ 0.0100          & 0.0395 $\pm$ 0.0085  & 0.0360 $\pm$ 0.0082   \\ \hline
\end{tabular}
\caption{Approximate worst-case defender utilities on the synthetic $15\times15$ grid at different checkpoints. The ``$\pm$'' indicates $95\%$
    confidence intervals over the 2000 testing episodes.}
\label{ckpt15}
\end{table*}

\begin{table*}
\centering
\begin{tabular}{l|ccccc}
\hline
Episodes      & 1e6 & 2e6 & 3e6 & 4e6 & 5e6 \\ \hline\hline
T=30  & 0.0070 $\pm$ 0.0036  & 0.1025 $\pm$ 0.0135  & \textbf{0.1225 $\pm$ 0.0140}  & 0.0740 $\pm$ 0.0115   & 0.0655 $\pm$ 0.0115  \\ \hline
T=300 & 0.0057 $\pm$ 0.0031   &  0.0210 $\pm$ 0.0062      & 0.0550 $\pm$ 0.0100   & 0.0560 $\pm$ 0.0101  & \textbf{0.0720 $\pm$ 0.0113}  \\ \hline
\end{tabular}
\caption{Approximate worst-case defender utilities on the extracted Singaproe road network at different checkpoints. The ``$\pm$'' indicates $95\%$
    confidence intervals over the 2000 testing episodes.}
\label{ckptsg}
\end{table*}

\begin{table*}
\centering
\begin{tabular}{l|c|c|c|c}
\hline 
Episodes      & NSG-NFSP & Uniform  Policy & Greedy Policy &IGRS++ \\ \hline
T=300 & \textbf{0.4350 $\pm$ 0.0217} &      0.0385  $\pm$ 0.0084     & 0 $\pm$ 0  &     OOM      \\ \hline 
\end{tabular}
\caption{Approximate worst-case defender utilities on the extracted Manhattan road network. The ``$\pm$'' indicates $95\%$
    confidence intervals over the 2000 testing episodes. OOM stands for Out of Memory.}
\label{mh}
\end{table*}
\section{More Results on Large-Scale NSGs}
\label{checkpoint}
We provide the approximate worst-case defender utilities in the synthetic road network and the extracted Singapore road network at different checkpoints in Table \ref{ckpt15} and Table \ref{ckptsg}, respectively.

We also conduct experiments on a road network extracted from the Manhattan island of New York. There are 704 nodes and 3114 edges in total. The initial locations of the four security resources and the attacker are chosen randomly, and five exit nodes are selected based on the connection to the outside world. We set the time horizon $T$ as 300. When do evaluation, the DQN attacker is trained for $1\times10^5$ episodes. We present the experimental results in Table \ref{mh}. It can be found that NSG-NFSP obtains superior performance over the baselines.

\begin{algorithm*}
\caption{NSG-NFSP}
\label{algo}
Create \textit{node2vec} embeddings $\mathcal{G}$ for nodes in the road network\;
Initialize modules of the defender, including the BR policy network $Q(s,a|\theta^{Q})$, the average policy network $\Pi(s,a|\theta^{\Pi})$, and the corresponding replay memories, $\mathcal{M}_{RL}$ (circular buffer) and $\mathcal{M}_{SL}$ (reservoir buffer)\;
Initialize modules of the attacker, including the MAB, the Cache, and the AVGer\;
\For{episode $\tau=1$ to $M$}{
Both players sample to set behaviour for the current episode, either as the BR policy ($Q(s,a|\theta^{Q})$, the MAB) or the average policy ($\Pi(s,a|\theta^{\Pi})$, the AVGer)\;
\If{the attacker behaves as the AVGer}{
The attacker samples to decide whether to explore\;
}
The attacker samples an exit $\zeta_{\tau}$ by following the distribution of its current behaviour (the MAB, the AVGer or Exploration)\;
The attacker samples an escaping path to the selected exit\;
The defender observes initial state $s_0^{def}$ and legal actions $\mathcal{A}_{def}(s_0^{def})$\;
\For{time step $t=0$ to $T$}{
The defender inputs the state $s_t^{def}$ and legal actions $\mathcal{G}(\mathcal{A}_{def}(s_t^{def}))$ into the network corresponding to its behaviour and samples an action $a_t^{def}$\;
Execute action $a_t^{def}$ in game and the defender observes reward $r_{t+1}^{def}$, state $s_{t+1}^{def}$ and legal actions $\mathcal{A}_{def}(s_{t+1}^{def})$\;
The defender stores transition $(s_t^{def},a_t^{def},r_{t+1}^{def},s_{t+1}^{def},\mathcal{A}_{def}(s_{t+1}^{def}))$ in $\mathcal{M}_{RL}$\;
\If{the defender is in the BR mode and the attacker is not in exploration}{
The defender stores tuple $(s_t^{def},a_t^{def})$ in $\mathcal{M}_{SL}$\;
}
}
Calculate attacker utility $u_{\tau}^{att}$ and the attacker stores tuple $(\zeta_{\tau},u_{\tau}^{att})$ in the MAB\;
\If{the attacker in the MAB mode}{
The attacker stores $\zeta_{\tau}$ in the Cache\;
}
The defender periodically updates $\theta^{Q}$ with SGD on the loss described in Eq. (\ref{loss_rl})\;
The defender periodically updates target network parameters $\theta^{Q'}\leftarrow\theta^{Q}$\;
The defender periodically updates $\theta^{\Pi}$ with SGD on the loss described in Eq. (\ref{loss_sl})\;
The attacker periodically updates the MAB by following Eq. (\ref{Q-MAB})\;
The attacker periodically updates the AVGer using data stored in the Cache and clear the Cache\;
}
\KwOut{The defender's average policy network $\Pi(s,a|\theta^{\Pi})$}
\end{algorithm*}
\end{appendices}

\end{document}